 \documentclass[final,3p,authoryear]{elsarticle}

 \makeatletter
\def\ps@pprintTitle{%
 \let\@oddhead\@empty
 \let\@evenhead\@empty
 \let\@oddfoot\@empty
 \let\@evenfoot\@empty
}
\makeatother

\usepackage[flushleft]{threeparttable}
\usepackage{subfigure}
\usepackage{footnote}
\usepackage{amsthm}
\usepackage{amsmath,amssymb}
\usepackage{graphicx,url}
\usepackage[usenames]{color}
\usepackage{booktabs,caption}
\usepackage{colortbl,dcolumn}
\usepackage{makecell}
\usepackage{tabularx}
\usepackage{multirow}
\usepackage{epstopdf}
\usepackage{svg}
\usepackage{booktabs}
\usepackage{multicol}
\usepackage{mathtools}
\usepackage[linesnumbered,ruled]{algorithm2e}

\usepackage{cite}
\biboptions{comma,sort&compress}

\DeclareMathOperator*{\argmax}{arg\,max}
\DeclareMathOperator*{\argmin}{arg\,min}

\begin{document}

\begin{frontmatter}

\title{A Knowledge-Informed Deep Learning Paradigm for Generalizable and Stability-Optimized Car-Following Models}

	\author[1]{Chengming Wang}
	\ead{Chengming.Wang23@student.xjtlu.edu.cn}

	\author[1]{Dongyao Jia\corref{cor1}}
 \cortext[cor1]{Corresponding author}

 \ead{Dongyao.Jia@xjtlu.edu.cn}

	\author[1]{Wei Wang}
	\ead{Wei.Wang03@xjtlu.edu.cn}
 
	\author[2]{Dong Ngoduy}
	\ead{Dong.Ngoduy@monash.edu}

	\author[3]{Bei Peng}
	\ead{Bei.Peng@liverpool.ac.uk}

	\author[4]{Jianping Wang}
	\ead{jianwang@cityu.edu.hk}

	\affiliation[1]{organization={School of Advanced Technology},
		addressline={Xi'an Jiaotong-Liverpool University}, 
		city={Suzhou},
		postcode={215123}, 
		country={China}}

    \affiliation[2]{organization={Institute of Transport Studies},
		addressline={Monash University}, 
		city={Clayton},
		postcode={3800}, 
		country={Austraila}}
  
      \affiliation[3]{organization={Department of Computer Science},
		addressline={University of Liverpool}, 
		city={Liverpool},
		postcode={L69 7ZX}, 
		country={UK}}

    \affiliation[4]{organization={Department of Computer Science},
		addressline={City University of Hong Kong}, 
		city={HongKong},
		country={China}}

\begin{abstract}

Car-following models (CFMs) are fundamental to traffic flow analysis and autonomous driving. Although calibrated physics-based and trained data-driven CFMs can replicate human driving behavior, their reliance on specific datasets limits generalization across diverse scenarios and reduces reliability in real-world deployment. Moreover, these models typically focus on behavioral fidelity and do not support the explicit optimization of local and string stability, which are increasingly important for the safe and efficient operation of autonomous vehicles (AVs). To address these limitations, we propose a Knowledge-Informed Deep Learning (KIDL) paradigm that distills the generalization capabilities of pre-trained Large Language Models (LLMs) into a lightweight and stability-aware neural architecture. LLMs are used to extract fundamental car-following knowledge beyond dataset-specific patterns, and this knowledge is transferred to a reliable, tractable, and computationally efficient model through knowledge distillation. KIDL also incorporates stability constraints directly into its training objective, ensuring that the resulting model not only emulates human-like behavior but also satisfies the local and string stability requirements essential for real-world AV deployment. We evaluate KIDL on the real-world NGSIM and HighD datasets, comparing its performance with representative physics-based, data-driven, and hybrid CFMs. Both empirical and theoretical results consistently demonstrate KIDL’s superior behavioral generalization and traffic flow stability, offering a robust and scalable solution for next-generation traffic systems.

\end{abstract}

\begin{keyword}

Car-following models\sep Large language models\sep Knowledge distillation\sep Stability Analysis\sep Deep learning
\end{keyword}

\end{frontmatter}

\section{Introduction}
\label{Introduction}

Car-following models (CFMs) are microscopic traffic models that capture longitudinal interactions between leading and following vehicles. They play a central role in Intelligent Transportation Systems (ITS), simulating how vehicles adjust speed and position in response to surrounding traffic. As traffic systems evolve to include mixed flows of human-driven vehicles (HDVs) and autonomous vehicles (AVs), CFMs that generalize across diverse conditions and ensure traffic flow stability are increasingly important for optimizing ITS and addressing emerging mobility challenges \citep{wangCarFollowingModelsHumanDriven2023}.

Most CFMs follow a model-centric design and are calibrated or trained on specific datasets. While this yields high accuracy within seen scenarios, performance often degrades under unseen conditions due to the out-of-distribution generalization problem \citep{wangGeneralizingUnseenDomains2022,liuOutDistributionGeneralizationSurvey2023}. Individual datasets rarely capture the full range of real-world variability, limiting model robustness. Although data-centric approaches that focus on collecting broader datasets can improve generalization, they are costly and difficult to scale. Moreover, since CFMs are often designed with a primary focus on behavioral fidelity, they do not explicitly incorporate mechanisms to support system-level optimization objectives such as local and string stability, which are essential for safe and efficient traffic flow in AV-integrated environments.

To address these emerging demands, we propose a Knowledge-Informed Deep Learning (KIDL) paradigm that jointly enhances behavioral generalization and traffic flow stability. 

KIDL improves generalization by distilling high-level car-following knowledge from large language models (LLMs), which are pre-trained on diverse textual sources including traffic regulations and driving manuals. This enables KIDL to capture principles that extend beyond the scope of any single dataset. Through knowledge distillation \citep{xuSurveyKnowledgeDistillation2024a}, insights from LLMs (as teachers) are transferred to lightweight neural networks (as students), forming a compact and efficient representation. Rather than employing LLMs as end-to-end models \citep{chenGenfollowerEnhancingCarfollowing2024, pengLCLLMExplainableLanechange2025}, KIDL adopts a distillation-based approach with three key advantages.

The first advantage is computational efficiency. LLMs generate linguistic responses sequentially and require substantial memory and processing resources, making them unsuitable for real-time applications \citep{kaddourChallengesApplicationsLarge2023}. In contrast, KIDL produces single-step numerical predictions with significantly fewer parameters, enabling real-time inference at a fraction of the computational cost. The second advantage is the prediction reliability. LLMs may produce inaccurate or unfaithful content \citep{jiSurveyHallucinationNatural2023}, which poses serious risks in safety-critical contexts. KIDL reduces this risk by applying self-consistency with majority voting during knowledge extraction \citep{wangSelfConsistencyImprovesChain2023}, improving reliability and minimizing the likelihood of erroneous behavior.

The third advantage is theoretical tractability. The black-box nature, complex architectures, and dependence on natural language inputs and outputs make LLMs difficult to interpret and unsuitable for formal analysis, limiting their applicability in stability studies such as local and string stability \citep{sunStabilityAnalysisMethods2018}. By distilling knowledge into a simplified surrogate model with numerical inputs and outputs, KIDL enables interpretable and analytically tractable stability analysis.

This property further allows KIDL to incorporate physically grounded stability constraints directly into the training objective, ensuring compliance with both local and string stability conditions. As a result, the model suppresses disturbance amplification and promotes smooth traffic flow. 

By integrating behavioral fidelity with stability optimization, KIDL provides a scalable and robust solution for deployment in mixed traffic environments. This combination of generalizable behavior modeling and explicit stability assurance addresses a critical gap between human driver emulation and control-oriented AV deployment. To the best of our knowledge, KIDL is among the first frameworks to systematically achieve both objectives within a unified paradigm by distilling car-following knowledge from LLMs into a stability-aware neural architecture.

Our contributions are as follows:
\begin{enumerate}
    \item We propose a knowledge-informed deep learning (KIDL) paradigm for developing a highly generalizable and theoretically stable CFM applicable in real-world mixed traffic contexts.
    \item We develop a lightweight deep neural network model to distill car-following knowledge from LLMs for generalization purposes, ensuring computational efficiency, prediction reliability, and theoretical tractability.
    \item We examine the local and string stability properties of LLM-based CFMs by using the KIDL model as a surrogate and implement constraints to optimize them for enhanced traffic flow stability.
    \item We conduct comprehensive experiments to validate the effectiveness of this paradigm, including evaluating the KIDL model's distillation performance, generalization capability across diverse traffic datasets, and conducting local and string stability analysis.
\end{enumerate}

The rest of the paper is organized as follows: Section \ref{sec-lr} offers an overview of related work, and Section \ref{sec-med} elaborates on our proposed method. Experimental results and analysis are presented in Section \ref{sec-exp}, followed by the conclusions in Section \ref{sec-cl}.

\section{Literature review}\label{sec-lr}

\subsection{Car-Following Models}

Car-following behaviors have been studied for over 90 years, with early work centered on physics-based car-following models (CFMs) \citep{bandoDynamicalModelTraffic1995,treiberCongestedTrafficStates2000}. These models use interpretable parameters, such as reaction time, desired speed, and desired headway \citep{parasharReassessingDesiredTime2025}, enabling theoretical analysis of traffic dynamics, including local and string stability \citep{treiberTrafficFlowDynamics2013}. Parameter estimation in physics-based CFMs, known as calibration, involves solving an optimization problem to minimize discrepancies between simulated and real-world trajectories. Recent advancements in calibration techniques include accounting for serial correlation \citep{zhangBayesianCalibrationIntelligent2024,zhangCalibratingCarfollowingModels2024}, intra-driver heterogeneity \citep{zhangGenerativeCarfollowingModel2022}, and feature-sharing methods \citep{wangNovelFeatureSharingAutoRegressive2024}.

While physics-based models offer simplicity and interpretability, their reliance on predefined rules limits adaptability to complex or heterogeneous traffic conditions. To overcome these limitations, data-driven approaches have emerged, leveraging empirical traffic data and deep learning to enhance predictive performance. Sequence models effectively capture temporal dynamics \citep{maSequenceSequenceLearning2020}, while graph-based models account for spatial interactions \citep{suGraphConvolutionNetworks2020}. However, the black-box nature of data-driven CFMs complicates theoretical analysis. Recent efforts have applied auto-differentiation to evaluate string stability in neural CFMs \citep{zhangStringStabilityNeural2024}, although such methods emphasize analysis rather than optimization. Moreover, data-driven models often face issues of overfitting and data inefficiency, limiting their practicality.

To address these challenges, hybrid models have been proposed \citep{moPhysicsInformedDeepLearning2021,gengPhysicsinformedTransformerModel2023}, integrating physics-based constraints into data-driven learning. These models treat physical principles as regularizers, guiding the training process. Other work has focused on learning optimal car-following relationships directly from data \citep{liDiscoveringOptimalRelationship2025}. By combining the interpretability of physics-based models with the flexibility of data-driven methods, hybrid approaches offer a more balanced solution.

Although these advancements enhance CFMs' capabilities, existing CFMs still face a common limitation: their reliance on specific traffic datasets. Physics-based CFMs require real-world driving trajectories for parameter calibration, while data-driven and hybrid models depend on traffic datasets for model training. Therefore, these models may struggle to generalize to unseen traffic scenarios, posing a critical challenge to their practical application. 

\subsection{Large Language Models}

Large Language Models (LLMs) have demonstrated exceptional performance in natural language processing (NLP), driving progress across a wide range of domains \citep{naveedComprehensiveOverviewLarge2024}. Built primarily on transformer architectures \citep{vaswaniAttentionAllYou2017}, LLMs are trained on large-scale corpora to predict sequential tokens, enabling them to generate coherent and context-aware text. With billions of parameters, these models capture complex linguistic patterns and encode extensive real-world knowledge.

Recently, LLMs have been increasingly applied to domain-specific areas, including transportation. In traffic forecasting, they excel at modeling complex spatio-temporal dynamics, capturing dependencies across time and space to predict traffic patterns \citep{zhangLargeLanguageModels2024}. In autonomous driving, LLMs leverage common-sense reasoning and broad knowledge to interpret diverse scenarios, anticipate risks, and support context-aware decision-making \citep{liKnowledgedrivenAutonomousDriving2023}. 

\subsubsection{LLM Basics}

LLMs are typically pre-trained from scratch using vast, diverse datasets. These models can be further divided into general-purpose and domain-specific LLMs \citep{naveedComprehensiveOverviewLarge2024}. General-purpose LLMs, such as ChatGPT \citep{openaiGPT4TechnicalReport2024}, are designed to perform a broad range of tasks, including language generation, translation, summarization, and question-answering. In contrast, domain-specific LLMs are trained on data from specific fields, providing them with specialized knowledge and language patterns customized to those areas. For instance, urban foundation models \citep{zhangUrbanFoundationModels2024} have been developed by training on extensive corpora of real-world urban data, enabling them to understand and predict complex urban dynamics, such as traffic flow.

The approach of using pre-trained LLMs can be broadly categorized into two types: Fine-tuning and Prompting.

Fine-tuning adapts pre-trained LLMs to specific tasks by further refining them on smaller, specialized datasets. This is essential in traffic-related tasks, which often involve complex dynamics that require task-specific adjustments. The instruction tuning paradigm has been introduced to enhance the predictive and reasoning capabilities of LLMs by training them on instruction-based traffic data \citep{xuDrivegpt4InterpretableEndend2024,liUrbanGPTSpatioTemporalLarge2024a,pengLCLLMExplainableLanechange2025}. 

Prompting is to provide LLMs with specific instructions or contexts to generate relevant outputs. For example, in traffic forecasting, prompting can be used to instruct the model to analyze traffic patterns over time \citep{guoExplainableTrafficFlow2024}. By providing prompts with detailed context, such as location-specific patterns or daily variations, LLMs can generate more accurate and context-sensitive traffic predictions. Additionally, by leveraging common-sense knowledge, pre-trained LLMs have shown promising results in vehicle control strategy with superior effectiveness \citep{cuiReceiveReasonReact2024,chenGenfollowerEnhancingCarfollowing2024}.

\subsubsection{LLM Challenges}\label{subsec-challanges}

While LLMs have demonstrated exceptional performance, they face three significant challenges when applied in real-world scenarios. The challenges can be summarized as computational cost, hallucinations, and intractability.

Addressing the challenge of computational cost has been a focus of ongoing research \citep{kaddourChallengesApplicationsLarge2023}, involving approaches like reducing LLM complexity and decoupling them from real-time tasks. Knowledge distillation, as extensively reviewed in \citep{xuSurveyKnowledgeDistillation2024a}, is a key technique that simplifies LLMs while retaining much of their functionality \citep{taveekitworachaiSpeedCostEffectiveLarge2024}. Another approach, asynchronous design \citep{chenAsynchronousLargeLanguage2025,shaLanguageMPCLargeLanguage2023a}, combines LLMs with real-time systems by using LLMs for long-term insights and faster models for real-time responses, mitigating the drawbacks of LLM complexity.

To address hallucination issues in LLMs, prompting methods like Chain-of-Thought (CoT) \citep{weiChainthoughtPromptingElicits2022} and self-consistency \citep{wangSelfConsistencyImprovesChain2023} have shown significant promise. CoT guides the model through step-by-step reasoning, breaking down complex tasks into smaller steps to improve accuracy and reduce errors by making the reasoning process more transparent. Self-consistency enhances reliability by generating multiple solutions to the same prompt and selecting the most consistent result through a majority vote mechanism. Additionally, carefully designed prompts can offer clear guidance and relevant context, effectively reducing hallucinations in LLM outputs.

Intractability, particularly in transportation applications, remains insufficiently addressed. LLMs’ complex architectures and reliance on natural language inputs hinder theoretical analysis and optimization, both of which are critical for ensuring safety and consistency in traffic systems. Without tractable formulations, it is challenging to guarantee compliance with core principles such as traffic flow stability or collision avoidance.

In summary, LLMs hold great promise for shaping modern transportation networks. However, significant challenges remain for their practical application, and many of these issues require further exploration.

\section{Methodology}\label{sec-med}

\begin{figure*}[!htbp]
  \centering
  \includegraphics[width=470pt, angle=0]{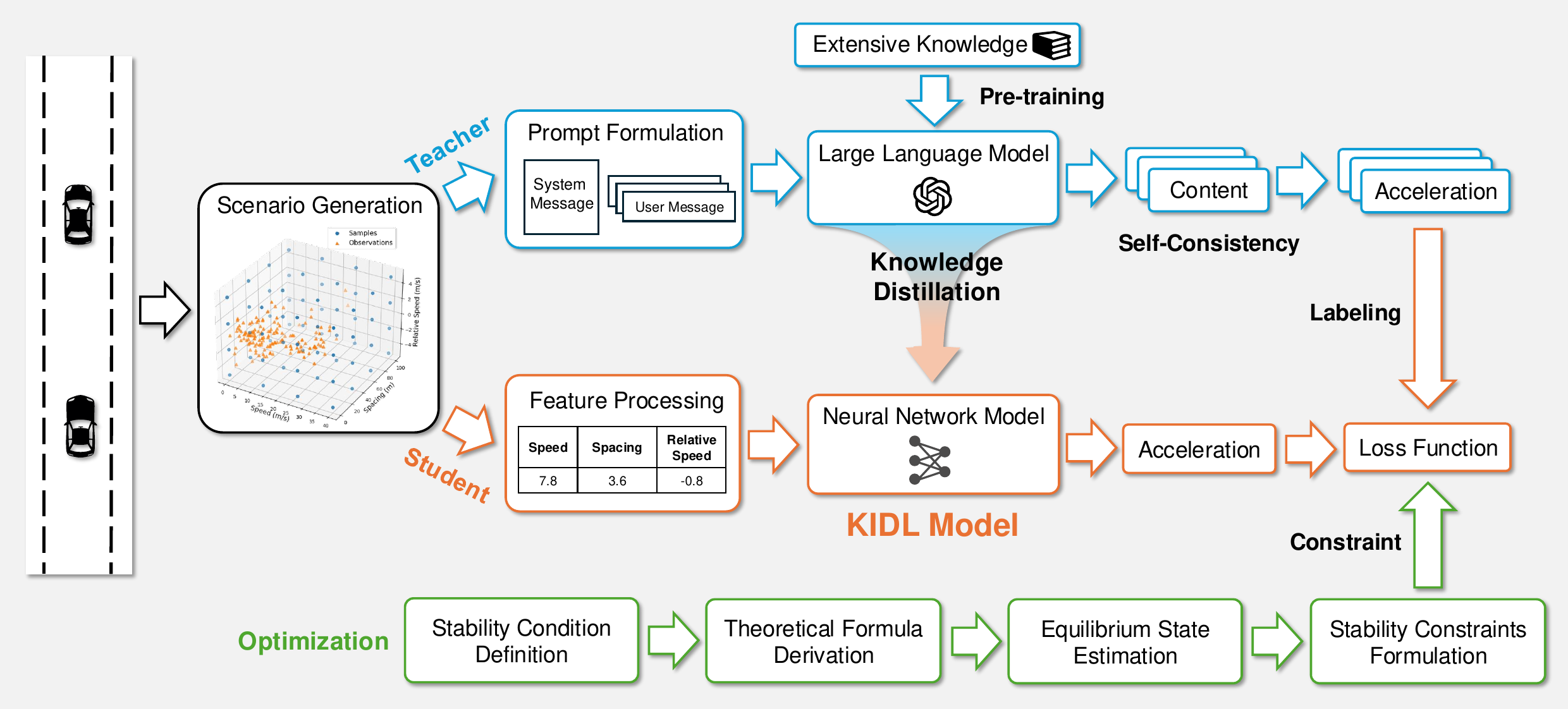}
  \caption{The knowledge-informed deep learning (KIDL) paradigm, with the blue section representing the LLM workflow (teacher demonstration), the orange section representing the distillation pipeline of KIDL (student learning), and the green section representing the stability optimization process.}
  \label{fig-framework}
\end{figure*}

Our proposed Knowledge-Informed Deep Learning (KIDL) framework for car-following modeling is illustrated in Figure~\ref{fig-framework}. The leftmost section depicts a typical car-following scenario, where the goal is to predict the acceleration of the following vehicle based on key variables: speed $v$, spacing $s$, and relative speed $\Delta v$. Orange triangular markers denote observations from real-world datasets, which cover only a limited subset of driving patterns. In contrast, blue circular markers represent the broader space of potential scenarios, highlighting the out-of-distribution (OOD) issue and motivating the use of LLMs to explore and generalize across the full scenario space.

The upper section, shown in blue, displays the workflow of the Large Language Model (LLM) as the teacher, where prompts are generated from the scenario variables within the full scenario space and fed into the LLM, which has been pre-trained on vast knowledge sources. The LLM produces reasoning contents from which accelerations are extracted. The middle section, shown in orange, displays the distillation pipeline of KIDL as the student, where the same scenario variables are processed as features and input into the deep neural network. The accelerations derived from the LLM serve as labels for training the neural network. The loss function measures the discrepancy between predicted and labeled accelerations, and the resulting loss is backpropagated to train the network. This process enables the distilled model to replicate the LLM’s acceleration prediction capabilities across diverse scenarios, achieving generalization comparable to or exceeding that of the LLM. 

Furthermore, the lower section, shown in green, displays the stability optimization process. Theoretical conditions derived from stability definitions are combined with the estimated equilibrium state to construct stability constraints, which are embedded into the distillation loss function. These constraints guide the KIDL model to achieve both behavioral generalization and traffic flow stability.

\subsection{Car-following Models}\label{subsec-cfm}

Car-following models (CFMs) describe the longitudinal interactions between a following vehicle and its leading vehicle. One primary goal of CFMs is to replicate human-driven vehicles' driving behaviors by defining a nonlinear function $f$ that maps the car-following state vector $S$ of the following vehicle to its corresponding action $a$:
\begin{equation}\label{eq-cfm}
a = f(S;\lambda)
\end{equation}

Here $\lambda$ represents the set of CFM parameters. The car-following state vector $S$ typically includes key variables such as the following vehicle's speed $v$, spacing with the leading vehicle $s$, and their relative speed $\Delta v$. These variables capture the essential components of car-following dynamics. The car-following action $a$ generally refers to the acceleration of the following vehicle in response to changes in the car-following state given the parameter set $\lambda$.

Using the mapping function, car-following trajectories can be simulated through ballistic update equations applied over intervals of $\Delta t$ seconds, where $x$ denotes the longitudinal position. These update equations, as shown in Equation \eqref{eq-ballistic}, enable the continuous simulation of vehicle trajectories based on the car-following state and corresponding accelerations.
\begin{equation}\label{eq-ballistic}
\begin{aligned}
v_{t+1} &= v_t + a_t\Delta t \\
x_{t+1} &= x_t + v_t\Delta t + \frac{1}{2} a_t\Delta t^2
\end{aligned}
\end{equation}

In addition to accurately replicating human driving behaviors, another primary goal of CFMs is to optimize these behaviors to ensure traffic flow stability. Stability, particularly local and string stability, is a critical property to enable vehicles to respond smoothly to small disturbances in speed or spacing while keeping equilibrium \citep{treiberTrafficFlowDynamics2013}. The equilibrium states refer to conditions in which speed and spacing are constant, denoted as $v_e$ and $s_e$, respectively. Assume the vehicles are ordered in a platoon, labeled from $1$ to $n$ in the upstream direction, with the $1$st vehicle acting as the leader. Disturbances in the equilibrium state can be expressed as follows:
\begin{equation}
\begin{aligned}
y_i(t) &= s_i(t) - s_e \\
u_i(t) &= v_i(t) - v_e
\end{aligned}
\end{equation}

where $y_i(t)$ and $u_i(t)$ denote the variations in spacing and speed of the $i$th vehicle at time $t$, and $s_i(t)$ and $v_i(t)$ represent the actual spacing and speed. The following analysis uses speed disturbances for demonstration.

Local stability examines how these variations evolve over time $t$, evaluating the magnitude of the disturbance for the following vehicle as time continues. A locally stable CFM can return to an equilibrium state after experiencing small perturbations, as shown in Equation \eqref{eq-ls}. In contrast, a locally unstable CFM amplifies these disturbances over time. This instability can lead to an increased risk of collisions and potential disruptions in traffic flow.
\begin{equation}\label{eq-ls}
\lim_{t\to\infty}|u_i(t)| = 0
\end{equation}

As outlined in \citep{treiberTrafficFlowDynamics2013,sunStabilityAnalysisMethods2018}, the general criterion for CFMs to ensure local stability is given by
\begin{equation}
f_v\big|_e + f_{\Delta v}\big|_e < 0 \ \text{and} \ f_s\big|_e > 0
\end{equation}

where $f_v\big|_e$, $f_s\big|_e$, and $f_{\Delta v}\big|_e$ are the Taylor expansion coefficients of the CFM function $f$ at the equilibrium states with respect to speed $v$, spacing $s$, and relative speed $\Delta v$. This criterion implies a rational human driving constraint that acceleration should have consistent, monotonic relationships with speed, spacing, and relative speed. For instance, increasing the spacing between vehicles generally leads to larger accelerations, assuming other conditions remain constant. Moreover, the monotonicity constraint can be generalized to all states, not limited to equilibrium conditions. The monotonicity criterion is given by:
\begin{equation}\label{eq-ls-c}
f_v < 0, \ f_s > 0 \ \text{and} \ f_{\Delta v} < 0
\end{equation}

Clearly, CFMs that satisfy this monotonicity inherently ensure local stability, making monotonicity a sufficient but not necessary condition.

String stability examines how disturbances evolve across a sequence of vehicles $n$, assessing their propagation within a vehicle platoon. A string-stable CFM ensures that these disturbances diminish as they pass through the vehicle stream, as shown in Equation \eqref{eq-ss}. In contrast, a string-unstable CFM amplifies disturbances as they propagate, which can lead to traffic waves or stop-and-go conditions. This not only increases the probability of accidents but also reduces overall traffic flow efficiency.
\begin{equation}\label{eq-ss}
\|u_1\|_\infty > \|u_2\|_\infty > \cdots > \|u_n\|_\infty
\end{equation}

where the notation $\|u_i\|_\infty = \max_t |u_i|$ represents the maximum disturbance magnitude of speed for the $i$th vehicle across all times.

As outlined in \citep{treiberTrafficFlowDynamics2013,sunStabilityAnalysisMethods2018}, the general criterion for CFMs to ensure string stability is given by
\begin{align}\label{eq-ss-c}
f_v^2\big|_e - 2f_s\big|_e + 2f_v\big|_ef_{\Delta v}\big|_e > 0
\end{align}

In summary, both monotonicity and string stability are closely tied to the partial derivatives of the CFM function with respect to scenario variables. Therefore, it is essential that newly developed CFMs facilitate accurate calculation of these partial derivatives to enable a precise assessment of stability conditions.

\subsection{Large Language Model for CFMs}

A major limitation of traditional CFMs is their reliance on real-world traffic datasets, which often cover a limited range of driving scenarios. Optimizing models on such constrained data can lead to overfitting and poor generalization to unseen conditions, thereby reducing their reliability in practical applications.

In contrast, Large Language Models (LLMs) are trained to infer patterns from diverse textual contexts, enabling them to model complex input–output relationships without task-specific supervision. When adapted to car-following modeling, LLMs treat driving as a sequence prediction task, reasoning over traffic states to generate plausible acceleration decisions. This allows them to generalize beyond the limited coverage of trajectory datasets and capture context-dependent driving behaviors \citep{chenGenfollowerEnhancingCarfollowing2024}.

In the LLM-based CFM, the car-following state vector $S$ is embedded into the input prompt via an embedding function $g(S)$, which conveys critical guidance and state information. The LLM processes this prompt with reasoning to infer the car-following scenario and outputs the predicted acceleration $a^*$. The mapping of LLM-based CFMs is defined as:
\begin{align}\label{eq-llm}
a^*, r^* = \argmax_{a_i\in \mathbb{A},r_i\in \mathbb{R}} P(a_i,r_i|g(S);\lambda)
\end{align}

Here, $a_i$ represents the acceleration prediction, selected from a fixed set $\mathbb{A}$ (e.g., a discretized range of acceleration values). $r_i$ represents a latent variable denoting the reasoning path that leads to the acceleration prediction $a_i$, belonging to a potential reasoning space $\mathbb{R}$. The parameters of the pre-trained LLM are denoted by $\lambda$. The LLM-based CFM uses a greedy decoding strategy, generating both the reasoning path $r^*$ and the corresponding acceleration prediction $a^*$ that maximize the probability $P$, conditioned on the car-following state vectors encapsulated within the prompts. 

The LLM-based CFM introduces a novel way of conceptualizing car-following behavior by shifting from data-dependent modeling to a knowledge-driven paradigm. However, despite this conceptual advance, directly deploying LLMs in car-following applications remains challenging due to their high computational cost, susceptibility to hallucinations, and difficulty in theoretical analysis and optimization. These limitations highlight the need for more efficient formulations that retain the generalization capabilities of LLMs while meeting the practical demands of the next-generation traffic applications.

\subsection{Knowledge-informed CFM}

To adapt LLM-based CFMs to real-world demands as identified above, we propose a Knowledge-Informed Deep Learning (KIDL) paradigm. This framework is based on knowledge distillation \citep{xuSurveyKnowledgeDistillation2024a}, where a large teacher model (LLM) transfers its learned knowledge to a smaller, more efficient student model. KIDL enables generalizable car-following behavior modeling while facilitating theoretical analysis and optimization, such as local and string stability.

The implementation of the KIDL paradigm comprises four key components: scenario generation, prompt formulation, model design, and stability optimization. Each component addresses a critical aspect of the framework: scenario generation ensures representative coverage of driving conditions; prompt formulation bridges traffic state representation and LLM interaction; model design focuses on constructing and training a compact student network; and stability optimization incorporates local and string stability constraints into the learning objective.

\subsubsection{Scenario Generation via Distributional Sampling}

Scenario generation defines synthetic car-following situations used to query the LLM teacher, producing human-like acceleration predictions that serve as supervision labels for training the KIDL model. The main challenge lies in creating a sufficiently diverse and representative scenario set, such that the student model can generalize well to unseen conditions after distillation.

Traditional CFMs rely on three core state variables: vehicle speed $v$, spacing $s$, and relative speed $\Delta v$, to describe leader-follower interactions. As these variables are continuous and bounded, a natural approach is to model their empirical distributions and perform random sampling from them to generate realistic and diverse scenarios.

\begin{figure*}[!htbp]
  \centering
  \includegraphics[width=400pt, angle=0]{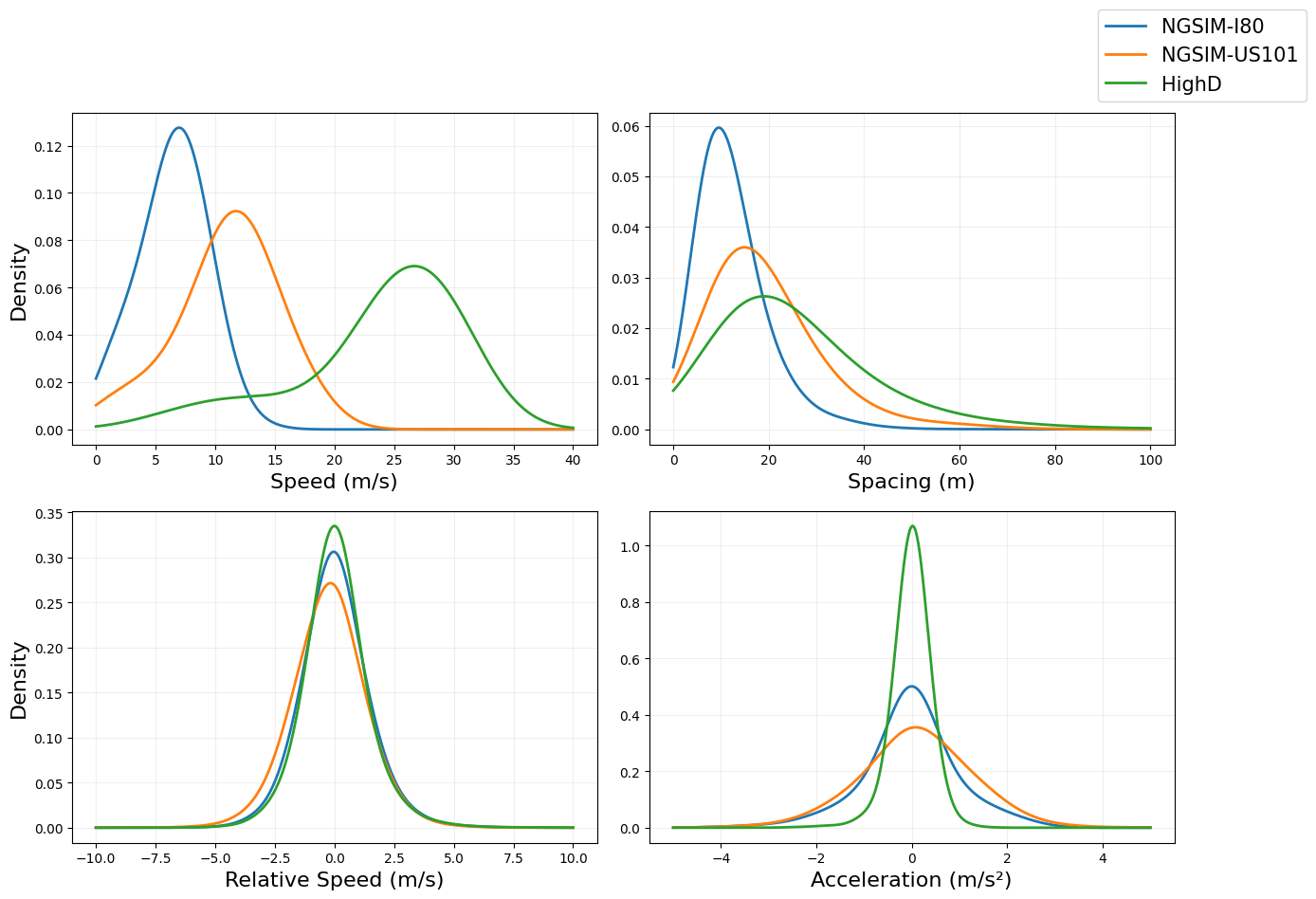}
  \caption{Distributions of four scenario variables, estimated via kernel density estimation, across the NGSIM-I80, NGSIM-US101, and HighD datasets.}
  \label{fig-dist}
\end{figure*}

To this end, we analyze the marginal distributions of scenario variables from three public trajectory datasets: NGSIM-I80, NGSIM-US101, and HighD \citep{krajewskiHighdDatasetDrone2018}. These datasets are used solely to estimate distributional characteristics, rather than to extract scenario instances. Figure~\ref{fig-dist} presents kernel density estimates (KDEs) of the scenario variables across the datasets. We observe that each variable approximately follows a truncated normal distribution, though the specific parameters vary across datasets due to differing traffic conditions.

For each scenario variable, the distribution patterns are generally consistent, except for the speed variable, which shows significant variations across datasets. This suggests that reducing the variable space could be a promising approach. To achieve this, we employ truncated normal distributions to model the distribution of scenario variables, thereby effectively reducing the variable space. Table \ref{tab-range} shows the truncated normal distribution parameters used for scenario generation in this study.
\begin{table}[!htbp]
\caption{The truncated normal distribution parameters for each scenario variable}
    \centering
    \begin{tabular}{lcccc} \toprule
\textbf{Variable} & \textbf{Mean} & \textbf{Std} & \textbf{Min} & \textbf{Max} \\ \midrule
Speed (m/s) & 15 & 15 & 0 & 40 \\ \addlinespace[0.5em]
Spacing (m) & 15 & 15 & 0.1 & 100 \\ \addlinespace[0.5em]
Relative Speed (m/s) & 0 & 2 & -5 & 5 \\ \bottomrule
\end{tabular}
    \label{tab-range}
\end{table}

\subsubsection{Prompt Formulation and Self-Consistency Decoding}

In traditional CFMs, scenario variables such as speed, spacing, and relative speed are used as features in parametric models to predict acceleration. In contrast, LLMs operate on natural language prompts, which are tokenized into numerical vectors to guide response generation.

To enable the LLM to generate structured, interpretable, and physically plausible acceleration predictions, we design a prompt that consists of two components: a system message that provides contextual background and a user message that defines the specific task. In the CFM context, the system message describes the car-following scenario and sets expectations for generating realistic, human-like behavior. The user message provides the scenario variable values for acceleration prediction.

\begin{figure*}[!htbp]
  \centering
  \includegraphics[width=400pt, angle=0]{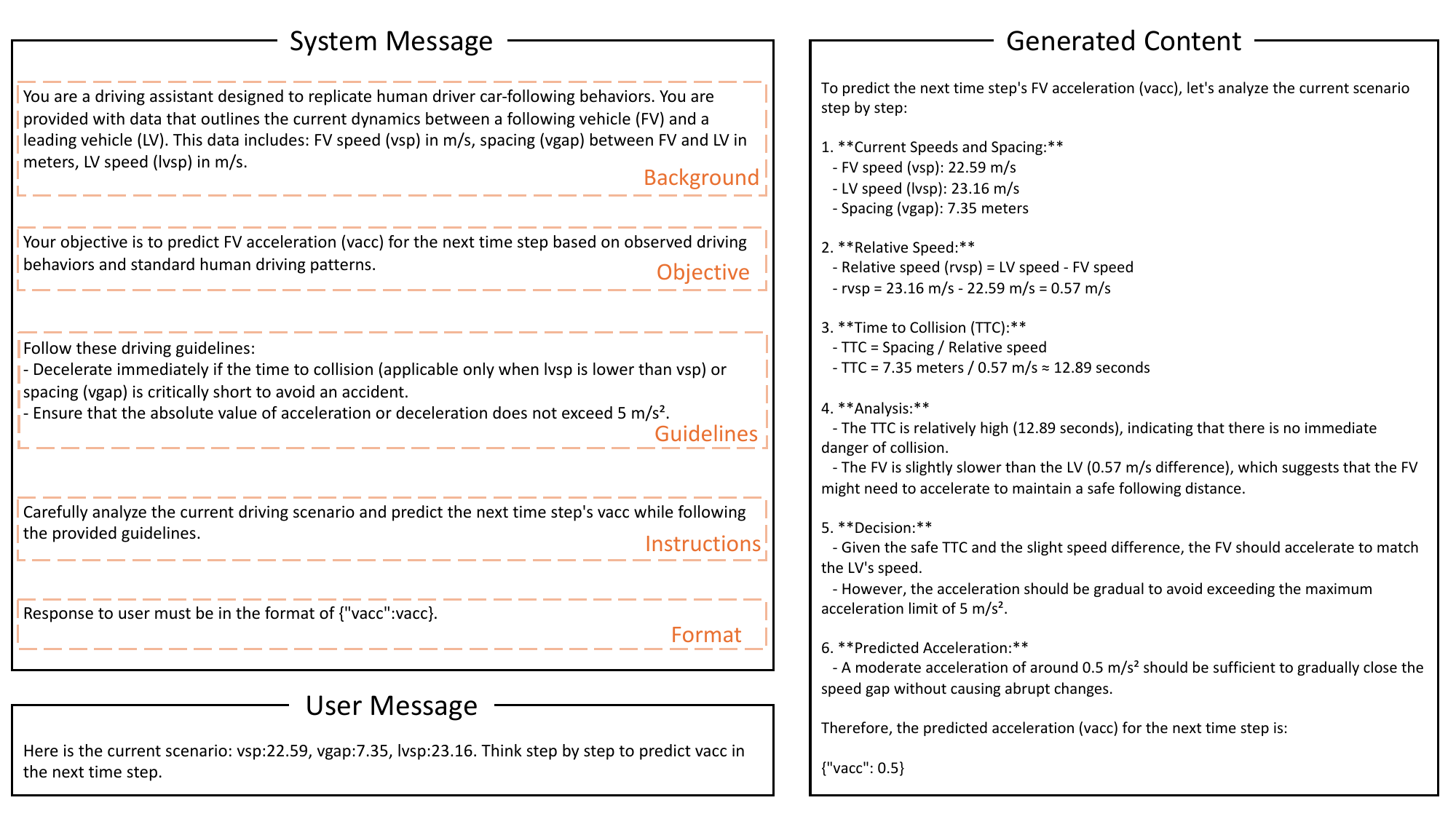}
  \caption{The system message used in this study, along with a user message and the generated content from the LLM.}
  \label{fig-prompt}
\end{figure*}

Figure \ref{fig-prompt} illustrates the system message adopted in this study, along with a sample user message and the corresponding LLM-generated output. The system message comprises five components: \textit{background}, \textit{objective}, \textit{guidelines}, \textit{instructions}, and \textit{format}. These components define the LLM's role, goal, operational constraints, reasoning steps, and output format, respectively.

The system message includes two key behavioral constraints: maintaining driving safety, and bounding acceleration values within plausible physical limits. Additional guidelines are avoided to preserve the generalizability of distilled knowledge and reduce potential bias. The user message embeds scenario variables into a template that leverages a chain-of-thought prompting strategy \citep{weiChainthoughtPromptingElicits2022}, guiding the LLM through structured reasoning. As shown in Figure \ref{fig-prompt}, this approach enables the model to decompose complex tasks into manageable steps, enhancing the accuracy and interpretability of its predictions.

\begin{figure*}[!htbp]
  \centering
  \includegraphics[width=400pt, angle=0]{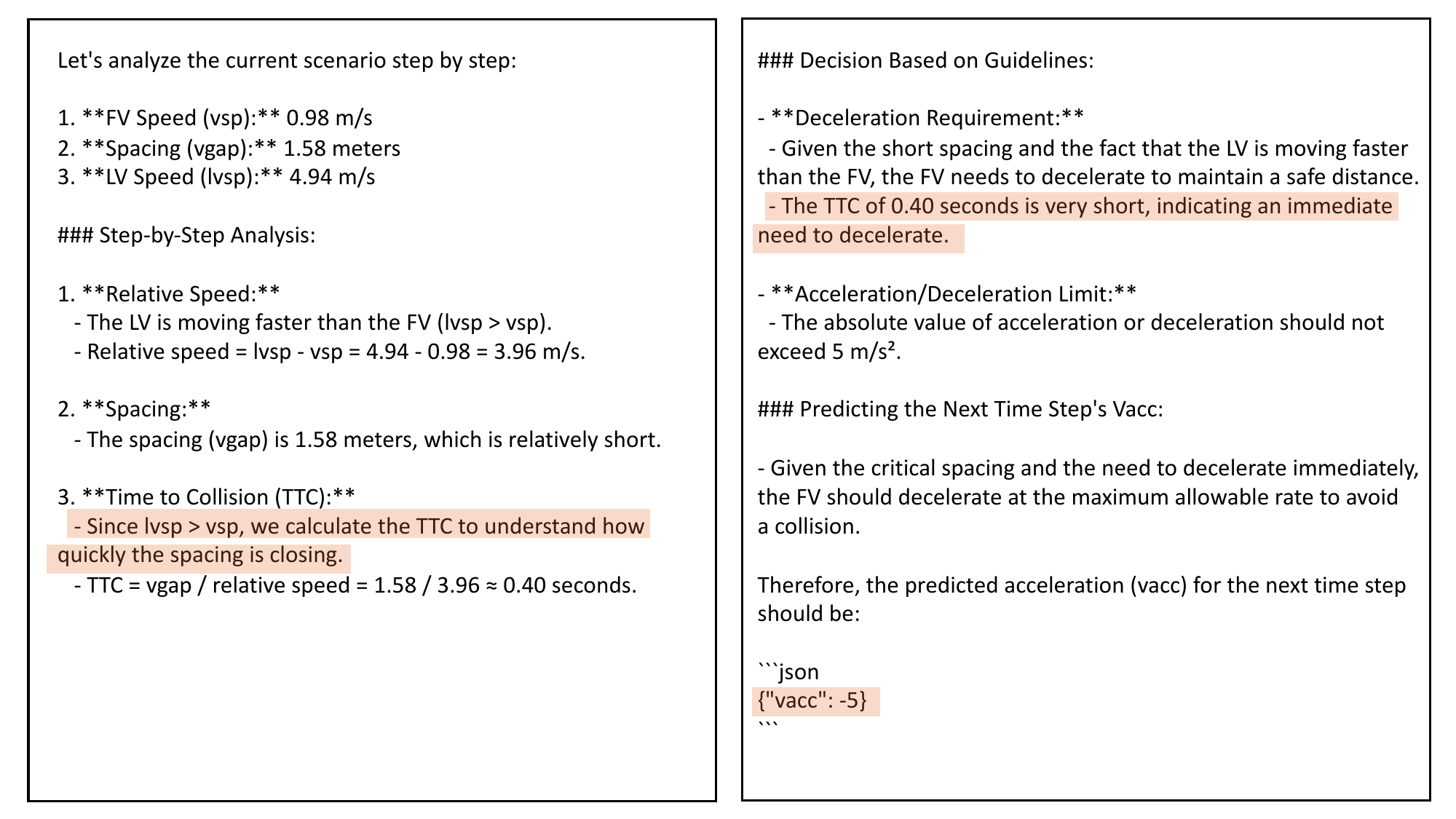}
  \caption{An example of hallucination generated by the LLM.}
  \label{fig-hal}
\end{figure*}

Despite the structured prompt formulation and reasoning guidance, LLMs may still produce erroneous outputs, commonly referred to as hallucinations \citep{jiSurveyHallucinationNatural2023}. In this context, hallucinations refer to predictions that deviate from plausible driving behavior or physical dynamics. Figure \ref{fig-hal} illustrates an example in which the LLM inaccurately computes the time-to-collision (TTC), resulting in an inappropriate sharp deceleration.

Such hallucinations stem from the discrepancy between LLMs' generative objective and the predictive objective of CFMs. Traditional CFMs directly map state variables to acceleration. In contrast, LLMs optimize for coherent linguistic responses, generating both reasoning and acceleration outputs. As formalized in Equation \eqref{eq-llm}, the LLM's mapping function does not inherently align with the goal of precise acceleration prediction. 

To bridge this gap, we propose a self-consistency-based reformulation of the prediction task \citep{wangSelfConsistencyImprovesChain2023}, leveraging the inherent probabilistic nature of LLMs. Instead of selecting the most probable single output, the model marginalizes over all possible reasoning paths $r_i$ to compute the conditional probability of each acceleration value: 
\begin{align}\label{eq-llmcfm}
a^* = \argmax_{a_i\in \mathbb{A}} P(a_i|g(S);\lambda) = \argmax_{a_i\in \mathbb{A}}\sum_{r_i\in \mathbb{R}}P(a_i, r_i|g(S);\lambda)
\end{align}

This formulation acknowledges that while the LLM may generate diverse reasoning paths, the objective remains to obtain reliable acceleration predictions. Marginalizing over these paths enhances robustness and better aligns the output with CFM objectives.

However, enumerating all reasoning paths is computationally infeasible. To address this, we approximate the marginalization by sampling a limited number of $(a_i, r_i)$ pairs and aggregating the acceleration predictions via majority voting.

In the example shown in Figure \ref{fig-hal}, with five iterations, only one produces this inappropriate acceleration of $-5 \ \text{m/s}^2$, while the other four iterations produce an appropriate acceleration of $1 \ \text{m/s}^2$. By applying a majority vote strategy, the model can effectively discard the outlier, thereby addressing the hallucination issue and ensuring a more accurate prediction.

\subsubsection{Model Design through Label-Based Distillation}\label{subsec-md}

The first two stages focus on teacher models, using the LLM to generate accurate human-like acceleration predictions from car-following scenarios. The third stage concentrates on student models (KIDL), which seeks to replicate the LLM's generalization capabilities in car-following modeling into a lightweight model for enhanced computational efficiency and theoretical tractability.

To maintain compatibility with the teacher, the student is implemented as a deep neural network, facilitating effective knowledge distillation. However, in contrast to the teacher, sequential architectures such as transformers are excluded, as the prediction task relies solely on the current car-following state and lacks temporal dependencies. Additionally, the student model operates on numerical inputs and outputs rather than natural language. These design choices significantly reduce model complexity and inference latency, enabling scalability for real-time and large-scale applications.

Given that many state-of-the-art LLMs (e.g., GPT-4 \citep{openaiGPT4TechnicalReport2024}) operate as black-box models, we adopt a labeling-based distillation strategy. This approach requires only the final acceleration prediction from the teacher model, which is treated as a ground-truth supervision signal for student training.

The student model is trained to minimize the discrepancy between its output and the teacher-provided labels. We adopt the mean squared error (MSE) as the loss function, which penalizes large deviations and encourages accurate replication of the LLM's predictions:
\begin{equation}
\begin{aligned}
\lambda^* &= \argmin_\lambda L_{\text{MSE}}(\hat{a}, f_{\text{student}}(S;\lambda)) \\
&= \argmin_\lambda \frac{1}{N} \sum_{i=1}^N (\hat{a}_i - f_{\text{student}}(S_i;\lambda))^2
\end{aligned}
\end{equation}

Here, $f_{\text{student}}$ denotes the mapping function of the student model, which predicts acceleration based on the car-following state $S_i$. The predicted value is compared against $\hat{a}_i$, the corresponding output from the LLM teacher. The optimized parameters $\lambda$ minimize the MSE loss over the training set.

\subsubsection{Stability-Constrained Optimization}\label{subsec-so}

The label-based distillation process enables the KIDL model to closely approximate the LLM's car-following behavior, resulting in a highly generalizable CFM. While such behavioral fidelity aligns well with the original design goals of CFMs, it does not inherently ensure critical system-level dynamical properties, such as local and string stability. These properties, which are essential for the safe and scalable deployment of AVs, are typically not addressed within the scope of behavior-driven model design.

To bridge this gap, we augment the KIDL model with stability analysis and optimization capabilities, leveraging its simplified, numerically defined structure. This structure allows for analytical gradient tracing from input state variables to acceleration outputs, Following the EADC framework \citep{zhangStringStabilityNeural2024}, stability assessment involves two steps: computing partial derivatives and estimating equilibrium states.

Partial derivatives are obtained via backpropagation using the chain rule across network layers. This method efficiently propagates gradients through the KIDL model, even in deeper architectures, yielding accurate derivatives with respect to each scenario variable.

Consider a deep neural network-based KIDL model with $L$ layers, where each layer's output is a non-linear transformation of the previous layer's outputs:
\begin{equation}
h^{(l)} = g^{(l)}(z^{(l)}) = g^{(l)}(W^{(l)}h^{(l-1)}+b^{(l)})
\end{equation}

Here, $h^{(l)}$ represents the output of the $l$th layer. $W^{(l)}$ and $b^{(l)}$ represent the weights and bias of that layer. $g^{(l)}$ is the activation function, which is typically non-linear. For each layer, the partial derivative of outputs with respect to inputs can be expressed as:
\begin{equation}
\frac{\partial h^{(l)}}{\partial h^{(l-1)}} = W^{(l)\intercal}\cdot\text{diag}(g^{(l)\prime}(z^{(l)}))
\end{equation}

where $\text{diag}(g^{(l)\prime}(z^{(l)}))$ is a diagonal matrix of the activation function's derivatives.

The partial derivative of acceleration predictions $\hat{a}$ with respect to scenario variables, such as spacing $s$, can then be calculated by applying the chain rule:
\begin{equation}
\frac{\partial \hat{a}}{\partial s} = \frac{\partial \hat{a}}{\partial h^{(L)}}\frac{\partial h^{(L)}}{\partial h^{(L-1)}}\cdots\frac{\partial h^{(1)}}{\partial s}
\end{equation}

The estimation of equilibrium states can be formulated as an optimization problem, as shown in Equation \eqref{eq-es}. It aims to determine the equilibrium speed and spacing that minimize the difference between the predicted acceleration and the equilibrium acceleration, which is ideally zero.
\begin{equation}\label{eq-es}
\begin{aligned}
\min_{v,s} |f(S;\lambda^*)&-0| \\
\text{subject to} \  \text{LB}_v &\leq v \leq \text{UB}_v \\
\text{LB}_s &\leq s \leq \text{UB}_s \\
\end{aligned}
\end{equation}

where $\text{LB}$ represents the lower bound and $\text{UB}$ represents the upper bound. The car-following state vector is defined as $S = (v, s, 0)$, where the relative speed equals $0$ in the equilibrium state. This optimization problem can be solved using a grid search over possible ranges of speed and spacing. However, if the CFM is not monotonic with respect to speed and spacing, multiple equilibrium solutions may emerge. Therefore, monotonicity for the entire scenario space is a prerequisite for accurate equilibrium state estimation.

To accomplish this, we propose enforcing monotonicity constraints across all samples, as shown in Equation \eqref{eq-mon-loss}. This approach imposes a stricter constraint than in the derivation of local stability in Equation \eqref{eq-ls-c}, where constraints are applied only at equilibrium states. Therefore, enforcing monotonicity across all samples inherently ensures local stability. The monotonicity constraint term $C_{\text{mon}}$ is expressed as
\begin{equation}\label{eq-mon-loss}
C_{\text{mon}} = \frac{1}{N}\left(\delta_v\sum_{i=1}^{N} \max\left(0, \frac{\partial \hat{a}_i}{\partial v_i}\right) + \delta_s\sum_{i=1}^{N} \max\left(0, -\frac{\partial \hat{a}_i}{\partial s_i}\right) + \delta_{\Delta v}\sum_{i=1}^{N} \max\left(0, \frac{\partial \hat{a}_i}{\partial \Delta v_i}\right)\right)
\end{equation}

where $\delta_v$, $\delta_s$, and $\delta_{\Delta v}$ are the penalty coefficients associated with speed, spacing, and relative speed, respectively. They control the penalty strength applied to enforce monotonicity for each variable.

For string stability, we first estimate equilibrium states with monotonicity enforced, then apply constraints based on the established string stability criterion, as shown in Equation \eqref{eq-str-loss}. The string stability constraint term $C_{\text{ss}}$ is expressed as
\begin{equation}\label{eq-str-loss}
C_{\text{str}} = \max\left(-\min_{i\in N_e}\left(\left(\frac{\partial\hat{a}_i}{\partial v_i}\right)^2 - 2\frac{\partial\hat{a}_i}{\partial s_i} + 2\frac{\partial\hat{a}_i}{\partial v_i}\frac{\partial\hat{a}_i}{\partial \Delta v_i} \right), 0\right)
\end{equation}

where the samples used for this calculation are all equilibrium states, with $N_e$ denoting their count. This loss penalizes the minimum value of string stability across all equilibrium states, ensuring larger string stability values above $0$ for all equilibrium states. 

These monotonicity and string stability constraints are incorporated directly into the student model’s loss function. By penalizing deviations from these desired properties, the model is encouraged to produce outputs that satisfy local and string stability. Moreover, monotonicity constraints enforce consistent acceleration responses aligned with rational human driving behavior, enhancing generalization beyond the capabilities of the LLM teacher model.

The final loss function of the KIDL student model is expressed as
\begin{equation}
L = L_{\text{MSE}} + \theta_{\text{mon}}C_{\text{mon}} + \theta_{\text{str}}C_{\text{str}}
\end{equation}

where $L_{\text{MSE}}$ denotes the distillation loss described in Section~\ref{subsec-md}, and $\theta_{\text{mon}}$ and $\theta_{\text{str}}$ represent the weighting factor for the monotonicity and string stability constraint, respectively. These weighting factors control the balance between replicating LLM predictions and optimizing local and string stability.

\section{Experiments}\label{sec-exp}

The experiments are structured into three parts. The first part focuses on assessing the distillation performance of the KIDL model to measure how well it inherits the car-following modeling capabilities of the LLM. The second part evaluates the generalization capability of the KIDL model by testing its performance across diverse driving scenarios. The third part demonstrates the stability optimization of the KIDL model, showing its superiority in obtaining theoretical insights into the LLM and achieving a more stable CFM through the KIDL paradigm.

\subsection{Datasets}
\begin{figure*}[!htbp]
  \centering
  \includegraphics[width=400pt, angle=0]{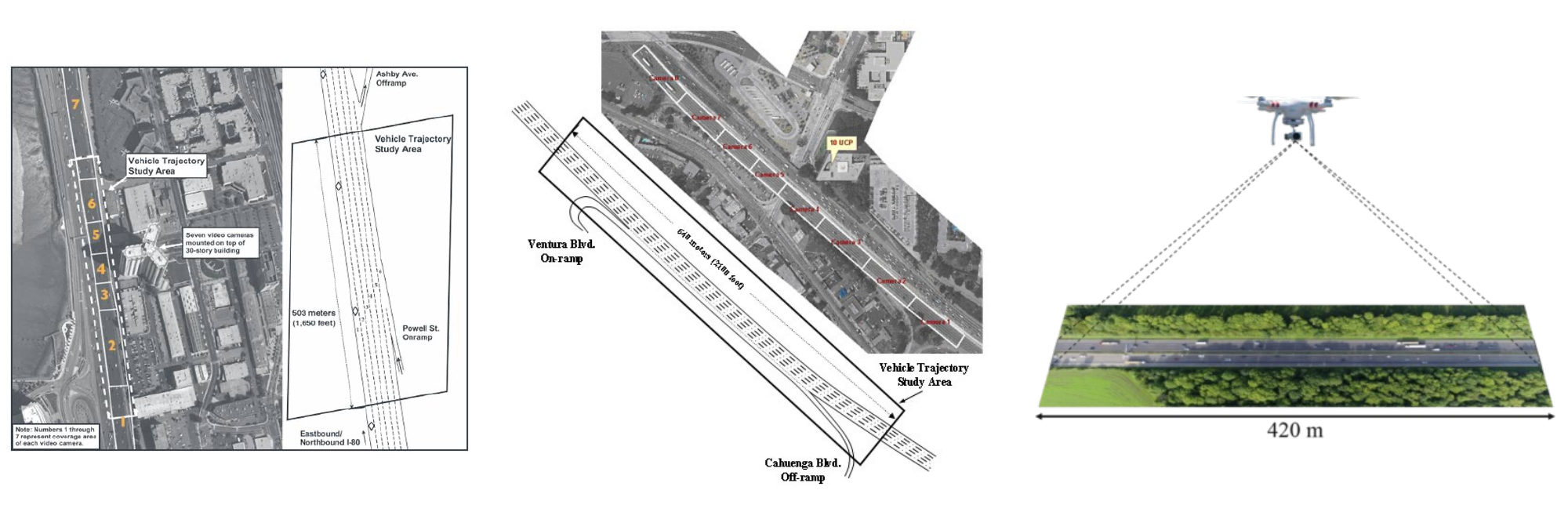}
  \caption{Left: NGSIM-I80, Middle: NGSIM-US101, Right: HighD}
  \label{fig-dataset}
\end{figure*}

Three of the most commonly used traffic datasets, NGSIM-I80, NGSIM-US101, and HighD, are included in this study to provide diverse and realistic car-following scenarios.
\begin{itemize}
    \item NGSIM-I80: As part of the Next Generation Simulation (NGSIM) project, this dataset contains vehicle trajectory data captured from a 500-meter segment of the I-80 freeway in Emeryville, California.

    \item NGSIM-US101: Also as part of the NGSIM project, this dataset covers a 640-meter segment of the US-101 freeway in Los Angeles, California. 
    
    \item HighD \citep{krajewskiHighdDatasetDrone2018}: This dataset was collected on German highways using drones, offering a bird's-eye view of vehicle trajectories over longer distances than the NGSIM datasets.
\end{itemize}

The NGSIM and HighD datasets are recorded at different frequencies: NGSIM at 10 Hz (data captured every 0.1 seconds) and HighD at 25 Hz (data captured every 0.04 seconds). Since CFMs are continuous models independent of sampling frequency, these datasets provide a platform to evaluate the model's adaptability to varying temporal resolutions.

These datasets are further processed to extract car-following trajectories \citep{montaninoTrajectoryDataReconstruction2015}. The detailed processing techniques include: 
\begin{itemize}
    \item Throughout the entire trajectory, the leading vehicle and lane position stay unchanged.
    
    \item To ensure driving stability, the trajectory duration is set to at least 30 seconds.

    \item The following and leading vehicles are restricted to automobiles, reducing variability from different vehicle types.
\end{itemize}

Trajectories in each dataset are partitioned into training, validation, and test sets using a 60-20-20 ratio. The training set is used for model training, while the validation set is used for parameter tuning and early stopping when the error curve converges during training. After training and validation, the test set is used for performance evaluation, providing an unbiased assessment of the model's effectiveness.

\subsection{Performance Metrics}

Acceleration prediction error is measured using the weighted mean absolute percentage error (WMAPE), which quantifies prediction accuracy by comparing the average absolute difference between predicted and actual accelerations to the average observed acceleration. WMAPE is chosen over the mean absolute percentage error (MAPE) because zero values in actual accelerations lead to an undefined denominator in MAPE. The term "weighted" refers to the weighting factor $p_i$, defined as the ratio of the actual acceleration of the  $i$th sample to the sum of all actual accelerations, as derived in the following formulation:
\begin{equation}
\begin{aligned}
\text{WMAPE} &= \frac{\sum_{i=1}^N |\hat{a}_i - a_i|}{\sum_{i=1}^Na_i} = \sum_{i=1}^N \left(\frac{a_i}{\sum_{i=1}^Na_i}\cdot\frac{|\hat{a}_i - a_i|}{a_i}\right) \\
             &= \sum_{i=1}^N \left(p_i \cdot \frac{|\hat{a}_i - a_i|}{a_i}\right)
\end{aligned}
\end{equation}

Trajectory simulation error is another performance metric, defined as the root mean squared error (RMSE) between the simulated trajectory spacing generated by the CFM and the actual trajectory spacing. Recommended by guidelines \citep{punzoCalibrationCarfollowingDynamics2021}, this metric evaluates the CFM's accuracy in replicating real-world vehicle trajectories. Lower RMSE values indicate better model performance in simulating the actual vehicle spacing. 
\begin{align}
\text{RMSE} &= \sqrt{\frac{1}{N}\sum\limits_{i=1}^N\sum\limits_{t=1}^T (s_{it} - \hat{s}_{it})^2}
\end{align}

Here, $\hat{s}_{it}$ is the predicted spacing for trajectory $i$ and time step $t$.

\subsection{CFMs for Comparison}

Five types of CFMs are evaluated for comparison, including three traditional models: physics-based, data-driven, and hybrid CFMs. To ensure fairness, all models are restricted to using the same three scenario variables as input features: speed $v$, spacing $s$, and relative speed $\Delta v$.
\begin{itemize}
    \item IDM \citep{treiberCongestedTrafficStates2000}: The Intelligent Driver Model (IDM) represents the physics-based CFM. It models desired driving behavior through interpretable parametric functions, with parameters calibrated using a genetic programming algorithm to best fit observed trajectories.

    \item DNN \citep{moPhysicsInformedDeepLearning2021}: A Deep Neural Network (DNN) serves as the data-driven baseline. It learns complex driving behaviors directly from trajectory data via layered nonlinear transformations.

    \item PIDL \citep{moPhysicsInformedDeepLearning2021}: The Physics-Informed Deep Learning (PIDL) model combines IDM and DNN. By embedding physics-based constraints into a deep learning framework, PIDL improves both accuracy and robustness in modeling car-following behavior.

    \item LLM \citep{chenGenfollowerEnhancingCarfollowing2024}: The Large Language Model (LLM), DeepSeek V2.5, is used to demonstrate the potential of LLM-based CFMs for its superior performance on open benchmarks and economical API access. Pre-trained on broad world knowledge, it captures realistic driving dynamics. Due to the high cost of API usage, the LLM is applied only in small-scale evaluations. Alternative LLMs are discussed in Appendix \ref{subsec-option}.

    \item KIDL: The Knowledge-Informed Deep Learning (KIDL) model distills knowledge from the LLM (DeepSeek V2.5) to achieve generalization and stability. A total of $10000$ car-following scenarios are generated, each queried from the LLM with $5$ iterations, yielding $50000$ samples. Majority voting determines the final label per scenario. The dataset is split into $80\%$ for training, $10\%$ for validation, and $10\%$ for testing. The distilled model is a deep neural network trained with monotonicity and string stability constraints. The monotonicity weight is set to $5000$, and the string stability weight to $0.9$. Monotonicity penalties are set to $0$ for speed and $1$ for both spacing and relative speed. All these parameter values are determined via grid search.
\end{itemize}

To comprehensively evaluate the effectiveness of each component of the KIDL paradigm, several ablation studies were conducted:
\begin{itemize}
    \item KIDL-basic: The simplest KIDL version, which predicts acceleration for all $50000$ samples with no added constraints. 
    \item KIDL-random: Based on KIDL-basic, it trains on $10000$ randomly selected samples, one per scenario. 
    \item KIDL-consist: Based on KIDL-basic, it trains on $10000$ samples chosen by majority vote for each scenario. This model serves as a surrogate for the LLM with self-consistency refinements.
    \item KIDL-mono: Based on KIDL-consist, it incorporates the monotonicity constraints into the training process to ensure local stability and further mitigate hallucinations of the LLM. 
\end{itemize}

\subsection{Distillation Performance}

This section evaluates KIDL's distillation performance in terms of acceleration prediction and trajectory simulation. For acceleration prediction, it compares KIDL's predicted accelerations with those from the LLM, assessing how well KIDL replicates the LLM’s acceleration prediction capability. 
\begin{table}[htbp]
\caption{Assessment of distillation performance based on acceleration prediction error}
    \centering
    \begin{tabular}{lcccc} \toprule
\textbf{Metric} & \textbf{KIDL-consist} & \textbf{KIDL-mono} & \textbf{KIDL} \\ \midrule
WMAPE & \textbf{0.206} & 0.297 & 0.337 \\ \bottomrule
\end{tabular}
    \label{tab-distill-acc}
\end{table}

Table $\ref{tab-distill-acc}$ presents the distillation performance measured by WMAPE. Among the KIDL models, KIDL-consist achieves the lowest error, with a WMAPE of $20.6\%$, demonstrating that the KIDL paradigm can closely replicate the LLM’s acceleration predictive capabilities. The higher errors observed for KIDL-mono and the full KIDL suggest that incorporating additional constraints reduces the LLM replication accuracy.

The second perspective concentrates on evaluating the KIDL model's distillation performance under real-world traffic conditions by comparing trajectory simulation errors between the LLM and the KIDL model. Here, the focus shifts from acceleration prediction to the CFM's ability to simulate realistic vehicle trajectories over time. However, due to the high API call costs associated with LLM usage, the duration of simulated trajectories is restricted to $15$ seconds, and the number of simulated trajectories is limited to $25$, which are randomly selected from the NGSIM-I80 dataset.
\begin{table}[htbp]
\caption{Assessment of distillation performance based on trajectory simulation error}
    \centering
    \begin{tabular}{lcccc} \toprule
\textbf{Metric} & \textbf{LLM} & \textbf{KIDL-consist} & \textbf{KIDL-mono} & \textbf{KIDL} \\ \midrule
RMSE & 4.060 & 4.161 & \textbf{3.761} & 3.784 \\ \bottomrule
\end{tabular}
    \label{tab-distill-tse}
\end{table}

Table $\ref{tab-distill-tse}$ presents a comparison of trajectory simulation error across four models: LLM, KIDL-consist, KIDL-mono, and KIDL. The results show that the KIDL paradigm effectively replicates the LLM’s trajectory simulation capabilities, as indicated by similar errors. Notably, KIDL-mono and KIDL outperform the LLM teacher model by incorporating monotonicity constraints, which encourage consistent driving behaviors and reduce hallucinations generated by the LLM. This suggests that exact replication of the LLM’s acceleration predictions may not be optimal for realistic trajectory simulation tasks, as LLMs may not adhere to rational human driving patterns, such as monotonicity.

\subsection{Generalization Performance}\label{subsec-gen}

This section assesses KIDL's generalization performance across various driving scenarios by comparing it with traditional CFMs using three distinct traffic datasets. The evaluation of traditional CFMs is conducted based on a cross-dataset approach, where each CFM is trained on one dataset and then tested across all three datasets. Errors are aggregated using a weighted average, with dataset weights based on the trajectory simulation errors of the IDM fitted on that dataset, referred to as IDM*. This approach tests the CFMs' ability to generalize beyond the specific conditions of the dataset they were trained on, as performing well across all three datasets indicates a strong generalization capability. 
\begin{table}[!htbp]
\caption{Trajectory simulation errors of IDM and KIDL across three datasets}
    \centering
    \begin{tabular}{llcccc} \toprule
\textbf{Train Data} & \textbf{Model} & \textbf{NGSIM-I80} & \textbf{NGSIM-US101} & \textbf{HighD} & \textbf{Aggregated} \\ \midrule
 & IDM* & 4.769 & 6.987 & 4.715 & 5.311 \\ \addlinespace[0.5em]
NGSIM-I80 & IDM & 4.769 & 7.972 & 6.466 & 6.218 \\ \addlinespace[0.5em]
LLM Samples & KIDL & 5.444 & 7.008 & 4.764 & 5.585 \\ \bottomrule
\end{tabular}
    \label{tab-demo}
\end{table}

Table \ref{tab-demo} presents the trajectory simulation errors for each of the three datasets, along with the aggregated errors. IDM* achieves the lowest error as it is trained and tested on the same dataset. To ensure a balanced comparison across datasets, the inverse of IDM* error on each dataset serves as a normalizer to standardize CFM errors across datasets. It is important to note that IDM* serves solely as the normalizer in this study and not as a baseline for evaluation, as it represents three separate models trained on individual datasets rather than a single model evaluated across all three datasets.

Table \ref{tab-demo} compares the performance of two models for demonstration purposes. IDM, trained on the NGSIM-I80 dataset, performs best on NGSIM-I80 but shows significant performance degradation when applied to other datasets, as evidenced by the higher errors on the NGSIM-US101 and HighD datasets. In contrast, KIDL, trained on LLM-generated samples without dependence on specific traffic datasets, demonstrates superior generalization, with an aggregated trajectory simulation error that is $10.18\%$ lower than IDM's. 
\begin{table}[!htbp]
\caption{Assessment of LLM generalization performance based on aggregated trajectory simulation error}
    \centering
    \begin{minipage}{0.45\textwidth}
        \centering
        \begin{tabular}{llcc} \toprule
        \textbf{Train Data} & \textbf{Model} & \textbf{Collision} & \textbf{Error} \\
        \midrule
        LLM samples & KIDL & 0 & \textbf{5.585} \\ \addlinespace[0.5em]
        LLM samples & KIDL-mono & 0 & 5.832 \\ \addlinespace[0.5em]
        NGSIM-I80 & IDM & 0 & 6.218 \\ \addlinespace[0.5em]
        LLM samples & KIDL-consist & 0 & 6.285 \\ \addlinespace[0.5em]
        LLM samples & KIDL-random & 0 & 6.325 \\ \addlinespace[0.5em]
        LLM samples & KIDL-basic & 0 & 6.366 \\ \addlinespace[0.5em] 
        NGSIM-US101 & PIDL & 76 & 7.371 \\ \bottomrule
        \end{tabular}
    \end{minipage}
    \hfill
    \begin{minipage}{0.45\textwidth}
        \centering
        \begin{tabular}{llcc} \toprule
        \textbf{Train Data} & \textbf{Model} & \textbf{Collision} & \textbf{Error} \\
        \midrule
        NGSIM-I80 & DNN & 149 & 7.709 \\ \addlinespace[0.5em]
        NGSIM-US101 & DNN & 127 & 8.110 \\ \addlinespace[0.5em]
        HighD & PIDL & 162 & 8.522 \\ \addlinespace[0.5em]
        NGSIM-I80 & PIDL & 108 & 9.294 \\ \addlinespace[0.5em]
        HighD & DNN & 249 & 11.377 \\ \addlinespace[0.5em]
        HighD & IDM & 0 & 11.934 \\ \addlinespace[0.5em]
        NGSIM-US101 & IDM & 0 & 13.600 \\ \bottomrule
        \end{tabular}
    \end{minipage}
    \label{tab-gen}
\end{table}

Table~\ref{tab-gen} reports the number of collisions and trajectory simulation errors for each CFM, aggregated across three traffic datasets. These CFMs are ranked by aggregated trajectory simulation errors and are organized into two tables, with training datasets specified for each. Notably, all KIDL variants rank highly, exhibiting low simulation errors and zero collisions, demonstrating the effectiveness of the KIDL paradigm.

Physics-based models, such as IDM trained on NGSIM-I80, also generalize well, achieving zero collisions and outperforming data-driven and hybrid models. Although some data-driven and hybrid models show competitive accuracy, they still produce collisions, limiting their practical applicability. Differences in recording frequency between HighD and NGSIM datasets contribute to performance degradation, particularly for CFMs trained on one dataset and evaluated on another. However, KIDL models are less affected by this degradation, indicating strong adaptability.

Among the KIDL variants, the full KIDL model ranks first, surpassing all traditional CFMs by at least $10.18\%$ in simulation accuracy. The poor performance of KIDL-basic, trained on all LLM outputs, suggests the presence of hallucination effects. KIDL-random, which filters out outliers through random sampling, offers marginal improvement. This is further enhanced in KIDL-consist, which applies majority voting to select consistent LLM outputs. KIDL-consist achieves performance comparable to the best traditional CFM, validating the effectiveness of LLM-derived knowledge.

However, LLM-based CFMs do not inherently satisfy key human behavioral properties such as monotonicity and string stability, limiting further gains. By incorporating monotonicity constraints, KIDL-mono reduces simulation error by $7.21\%$ relative to KIDL-consist. Adding string stability constraints further reduces error by $4.24\%$, highlighting the importance of integrating human behavioral principles into the distillation process.

\subsection{Stability Analysis}

In addition to generalization performance, stability is another crucial factor that ensures the developed CFMs can improve the safety and efficiency of the traffic flow when deployed in autonomous driving systems. However, the structure of LLMs poses significant challenges for stability analysis and optimization due to their complex structure with natural language inputs and outputs, which complicate the calculation of gradient flow. The KIDL paradigm solves this problem by directly processing numerical inputs and producing numerical outputs with a simplistic structure. 

\begin{figure*}[!htbp]
  \centering
  \includegraphics[width=400pt, angle=0]{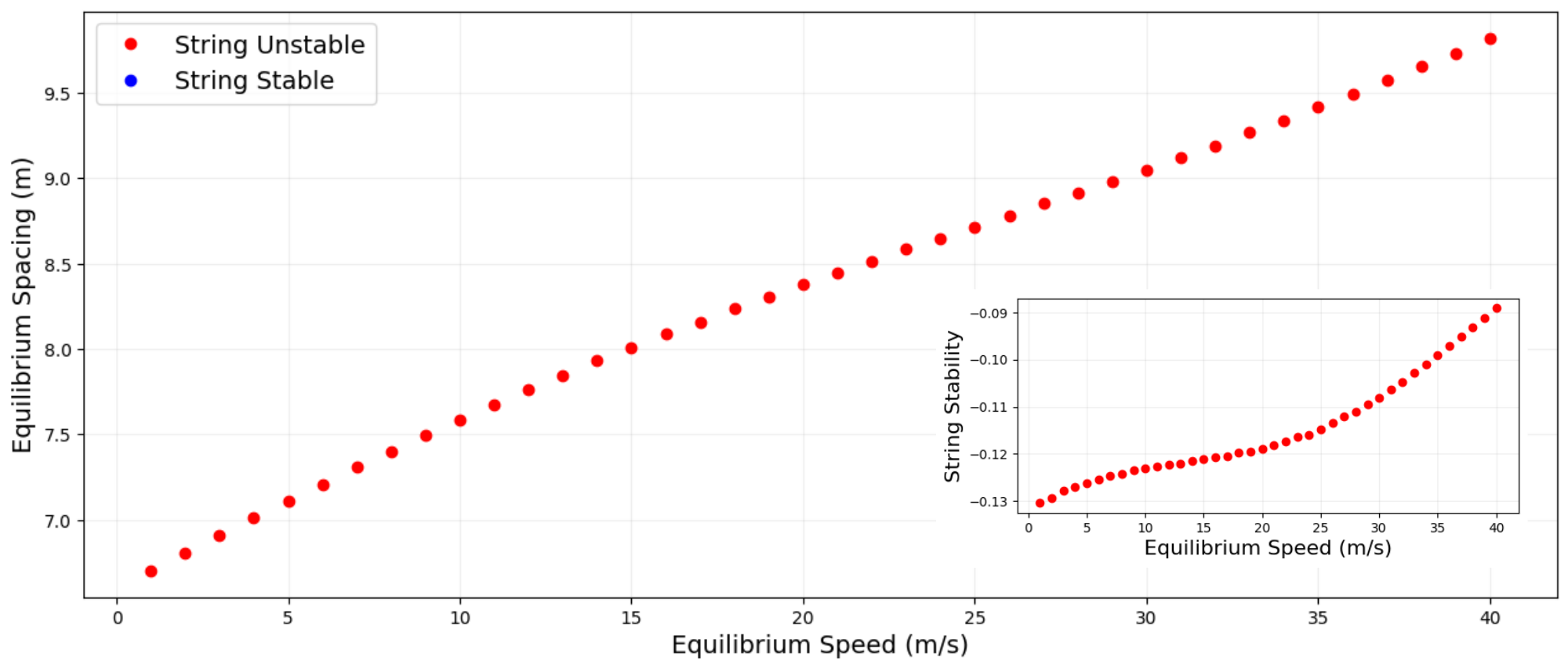}
  \caption{String stability analysis of the KIDL-mono model reveals that it is string unstable for all equilibrium states.}
  \label{fig-stb-ori}
\end{figure*}

The monotonicity constraint in the KIDL-mono model guarantees local stability across all equilibrium states. However, as illustrated in Figure \ref{fig-stb-ori}, the model remains string unstable at all equilibrium points. String stability results are computed for each equilibrium state using the derivation outlined in Section \ref{subsec-md}.

The results in Figure \ref{fig-stb-ori} reveal potential risks in deploying LLM-based CFMs, particularly in AV applications. String instability can degrade traffic flow efficiency and increase the likelihood of accidents, posing challenges for real-world implementation.

To mitigate this issue, a string stability constraint is integrated into the loss function of the KIDL-mono model, regulated by a weighting factor. The enhanced model is referred to as the full KIDL model.
\begin{figure*}[!htbp]
  \centering
  \includegraphics[width=400pt, angle=0]{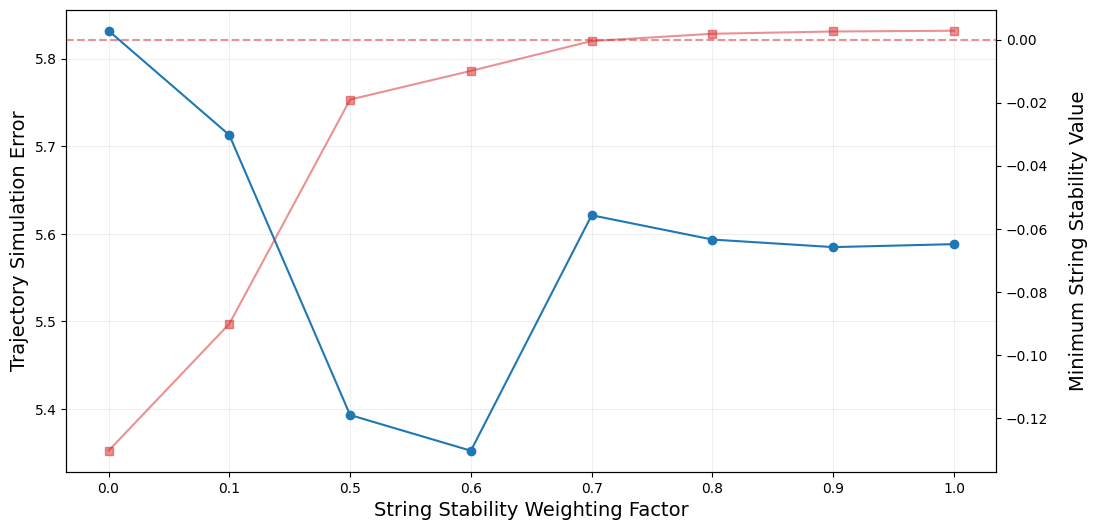}
  \caption{Performance evaluation of varying string stability weighting factors}
  \label{fig-comp-stb}
\end{figure*}

Figure \ref{fig-comp-stb} illustrates the effect of the string stability constraint weighting factor on generalization performance (blue, left axis) and string stability (red, right axis). String stability is measured by the minimum value of the left-hand side of Equation \ref{eq-ls-c}, evaluated across all equilibrium states. As the weighting factor increases, simulation error first decreases and then rises, while the string stability measure increases monotonically, eventually becoming positive, indicating stability.

The lowest simulation error occurs at a weighting factor of $0.6$, but the corresponding string stability value remains negative, indicating instability. To balance performance and stability, a weighting factor of $0.9$ is selected, ensuring string stability across all equilibrium states while maintaining low simulation error. These results suggest a trade-off between fidelity to human driving behavior and strict adherence to string stability, consistent with the empirical observation that real-world traffic is often string unstable.
\begin{figure*}[!htbp]
  \centering
  \includegraphics[width=400pt, angle=0]{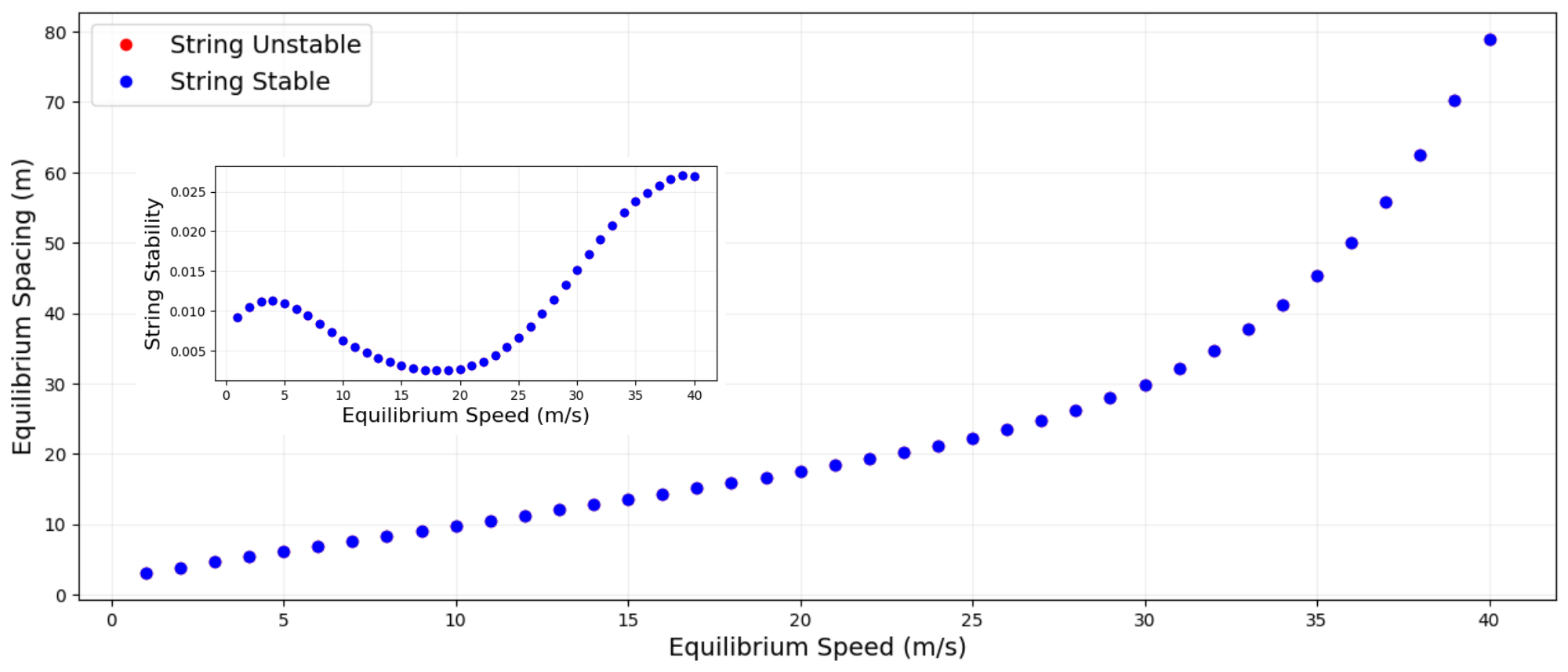}
  \caption{String stability analysis of the KIDL model reveals that it is string stable for all equilibrium states.}
  \label{fig-stb-opt}
\end{figure*}

Figure~\ref{fig-stb-opt} presents the string stability analysis of the full KIDL model with a weighting factor of $0.9$, confirming string stability across all equilibrium states. This demonstrates the effectiveness of the string stability constraint in enhancing overall model stability and promoting smoother, safer traffic flow. Additionally, as shown in Table \ref{tab-gen}, the full KIDL model also outperforms KIDL-mono in generalization, indicating that the constraint aids in capturing realistic human driving behavior. As a result, the KIDL model is both generalizable and stable, making it suitable for practical AV deployment.

To validate these theoretical findings, a numerical simulation is conducted following the setup in \citet{zhangStringStabilityNeural2024}. A homogeneous platoon of $100$ vehicles is simulated on a single-lane road using the KIDL model. Equilibrium speeds and spacing are predefined, and the simulation runs for $100$ seconds with a $0.1$ second time step. At $t = 6$ seconds, a disturbance is introduced: the lead vehicle decelerates at $0.5 \ \text{m/s}^2$ for $3$ seconds, followed by acceleration at the same rate for another $3$ seconds, before returning to equilibrium. The responses of the remaining $99$ vehicles are recorded to assess disturbance propagation. According to Equation \ref{eq-ss}, string stability is confirmed only if the disturbance diminishes monotonically along the platoon.
\begin{figure*}[!htbp]
  \centering
  \includegraphics[width=400pt, angle=0]{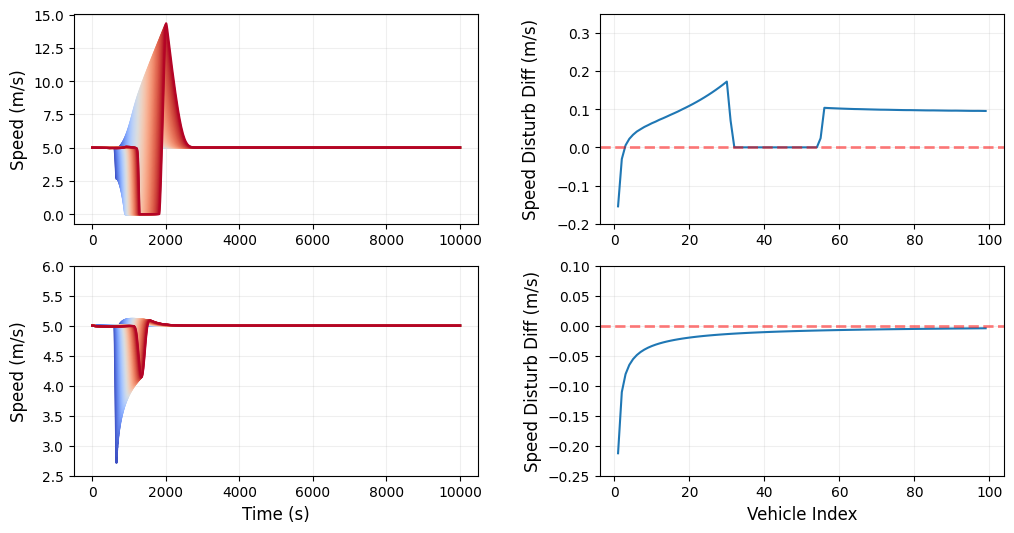}
  \caption{Numerical simulation results of the KIDL models. Upper: KIDL-mono. Lower: KIDL. The left side shows the speed changes over time, while the right side illustrates the speed disturbance differences between consecutive vehicles along the platoon.}
  \label{fig-num-stb}
\end{figure*}

Figure \ref{fig-num-stb} illustrates the propagation of speed disturbances through all following vehicles, represented as the speed changes and disturbance differences at an equilibrium speed of $5 \ \text{m/s}^2$ for demonstration purposes. The speed disturbance differences, defined as $\|u_i\|_\infty-\|u_{i-1}\|_\infty$, must remain non-positive to ensure string stability. As shown, the KIDL-mono model on the top exhibits string instability, whereas the KIDL model below exhibits string stability. These results are consistent with the findings from the theoretical analysis.

\section{Conclusion}\label{sec-cl}

In this study, we proposed a novel Knowledge-Informed Deep Learning (KIDL) paradigm that, to the best of our knowledge, is the first to unify behavioral generalization and traffic flow stability by systematically integrating high-level knowledge distillation from LLMs with physically grounded stability constraints in car-following modeling. Generalization is enhanced by distilling car-following knowledge from LLMs into a lightweight and efficient neural network, while local and string stability are achieved by embedding physically grounded constraints into the distillation process. Experimental results on real-world traffic datasets validate the effectiveness of the KIDL paradigm, showing its ability to replicate and even surpass the LLM's generalization performance. It also outperforms traditional physics-based, data-driven, and hybrid CFMs by at least $10.18\%$ in terms of trajectory simulation error RMSE. Furthermore, the resulting KIDL model is proven through theoretical and numerical analysis to ensure local and string stability at all equilibrium states, offering a strong foundation for advancing AV technologies.

Although the proposed KIDL framework shows strong foundational effectiveness, further improvements in fidelity and adaptability remain possible. Three key directions are outlined below.

First, enriching scenario representations with more contextual information. The current formulation defines car-following using speed, spacing, and relative speed, but omits other contextual information such as spatial-temporal awareness and memory effects. For instance, incorporating historical sequences can introduce memory effects, allowing KIDL to capture more consistent, human-like behavior and reduce hallucinations.

Second, incorporating intra- and inter-driver heterogeneity in prompt formulation. This study adopts a generalized human behavior model without addressing variability across drivers and contexts. Future work could incorporate driver preferences and environmental conditions to improve adaptability. For example, specifying an aggressive driver in a fast-lane scenario may yield shorter headways and higher speeds.

Third, improving LLM selection and domain alignment. KIDL’s performance depends on the quality of the LLM used for distillation. Using more advanced or traffic-specific LLMs can improve fidelity and reduce hallucinations. Fine-tuning general-purpose LLMs on real-world traffic data offers a promising path for modeling nuanced driving behaviors \citep{chenGenfollowerEnhancingCarfollowing2024, pengLCLLMExplainableLanechange2025}.

\section{Appendix}

\subsection{LLM Options}\label{subsec-option}

In this study, the KIDL paradigm is validated using distillation from the DeepSeek V2.5 model. A natural question arises as to whether this paradigm applies to other LLMs as well. To explore this, we conduct experiments on four additional popular LLMs:
\begin{itemize}
    \item GPT-4o: A more robust version of GPT-4 \citep{openaiGPT4TechnicalReport2024}, GPT-4o enhances reasoning and comprehension capabilities across complex tasks.
    \item GPT-4o-mini: An optimized variant of OpenAI's GPT-4 \citep{openaiGPT4TechnicalReport2024}, GPT-4o-mini is designed to reduce computational load while maintaining high performance. 
    \item Qwen-Max: Developed by Alibaba Cloud, Qwen-Max is a powerful LLM designed to handle highly complex tasks. It has demonstrated performance comparable to GPT-4o \citep{yangQwen2TechnicalReport2024}.
    \item Qwen-Plus: Also from Alibaba Cloud, Qwen-Plus is an efficient LLM tailored for tasks of medium complexity, offering a balance between performance and computational efficiency.
\end{itemize}
\begin{figure*}[!htbp]
  \centering
  \includegraphics[width=350pt, angle=0]{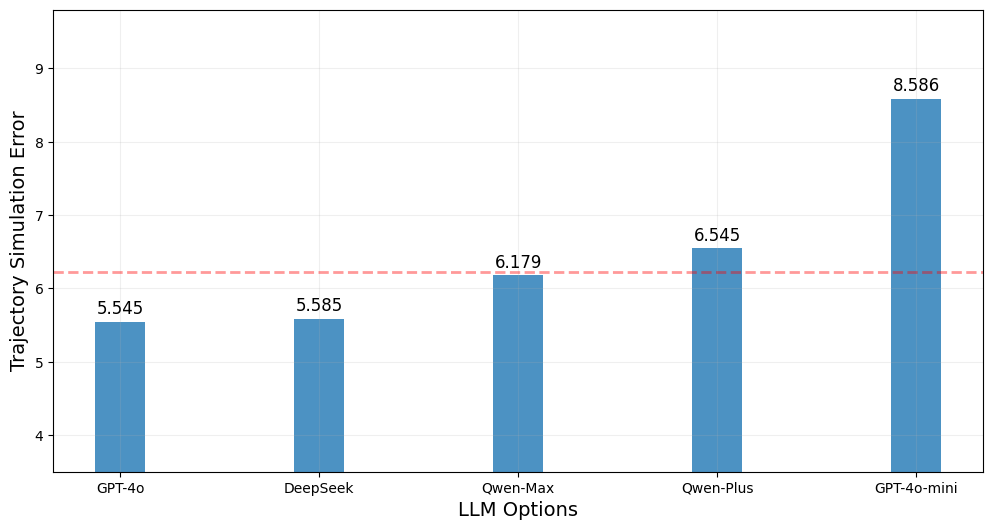}
  \caption{The aggregated trajectory simulation errors of the KIDL models distilled from different LLMs are presented, with a red dashed line indicating the error achieved by the best traditional CFM, the IDM model trained on the NGSIM-I80 dataset.}
  \label{fig-comp-llm}
\end{figure*}

Figure \ref{fig-comp-llm} presents the aggregated trajectory simulation errors of KIDL models distilled from different LLMs. Among them, the model distilled from GPT-4o achieves the lowest error, closely followed by DeepSeek V2.5. In contrast, GPT-4o-mini, a lightweight variant optimized for efficiency, yields the highest error. A similar trend is observed between Qwen-Max and Qwen-Plus, where the larger and more complex Qwen-Max not only outperforms traditional CFMs but also generalizes better than its smaller counterpart. These results indicate that the KIDL paradigm is broadly compatible with a range of LLMs, and its performance is expected to improve as more advanced LLMs become available.

\section*{Acknowledgments}

\end{document}